\documentclass[12pt]{article}
\usepackage{amsmath}
\usepackage{times}
\usepackage{color}
\usepackage{multirow}
\usepackage{url}
\usepackage{algorithm}
\usepackage{algpseudocode}
\usepackage[round]{natbib}
\usepackage{setspace}
\usepackage{amsfonts, mathtools, amssymb}
\usepackage{rotating}
\usepackage{bbm}
\usepackage{latexsym}

\newcommand{\captionfonts}{\normalsize}

\makeatletter  
\long\def\@makecaption#1#2{%
  \vskip\abovecaptionskip
  \sbox\@tempboxa{{\captionfonts #1: #2}}%
  \ifdim \wd\@tempboxa >\hsize
    {\captionfonts #1: #2\par}
  \else
    \hbox to\hsize{\hfil\box\@tempboxa\hfil}%
  \fi
  \vskip\belowcaptionskip}
\makeatother   

\begin{document}
\hspace{13.9cm}

\ \vspace{20mm}

\begin{center}
{\LARGE Closed-loop Multi-step Planning}
\end{center}

\ \\{\bf \large Lafratta, G.$^{\displaystyle 1}$}\\
{\bf \large Porr, B.$^{\displaystyle 1}$}\\
{\bf \large Chandler, C.$^{\displaystyle 2}$}\\
{\bf \large Miller, A.$^{\displaystyle 2}$}\\
{$^{\displaystyle 1}$School of Engineering, University of Glasgow}\\
{$^{\displaystyle 2}$School of Computing Science, University of Glasgow.}\\
%


\thispagestyle{empty}
\markboth{}{NC instructions}
\ \vspace{-0mm}\\

\begin{center} {\bf Abstract} \end{center}
Living organisms interact with their surroundings in a closed-loop fashion, where sensory inputs dictate the initiation and termination of behaviours. Even simple animals are able to develop and execute complex plans, which has not yet been replicated in robotics using pure closed-loop input control.
We propose a solution to this problem by defining a set of discrete and temporary closed-loop controllers, called ``Tasks'', each representing a closed-loop behaviour. We further introduce a supervisory module which has an innate understanding of physics and causality, through which it can simulate the execution of Task sequences over time and store the results in a model of the environment. On the basis of this model, plans can be made by chaining temporary closed-loop controllers. Our proposed framework was implemented for a real robot and tested in two scenarios as proof of concept.

\section{Introduction\label{intro}}
Living organisms interact with their surroundings through sensory inputs in a closed-loop fashion \citep{Maturana1980,Porr2005}. 
To recreate basic closed-loop navigation in a robot it is sufficient to directly connect a robot's sensors to its motor effectors, and the specific excitatory or inhibitory connections determine the control strategy, as in~\citet{Braitenberg1986}. These sensor-effector connections represent a single closed-loop controller, which produces control behaviours representing a response to an immediate stimulus.

Algorithms for autonomous navigation tend to fall into the categories of closed-loop output control, map-based, and trajectory planning algorithms.
Closed-loop output control in the form of Reinforcement Learning (RL) has been very successful in recent years, when supplemented with techniques such as deep learning~\citep{Mnih2015, Silver2016, OpenAI2023}. In RL, an agent uses environmental feedback to learn to output an optimal action in response to an input state, aiming to maximise the cumulative reward~\citep{Sun2021} where the state-space is typically discretised as a grid.
Environmental feedback about the ``value'' of an action can only be obtained \textit{retrospectively} at the end of a training episode. To avoid convergence to the local minimum, the action selection process is intentionally injected with random noise to varying degrees in order to encourage state-space exploration. These aspects make RL models inherently slow to train, to the point that an autonomous agent is not able to acquire knowledge tailored to its own environment in real time. Instead, large models must be trained offline in order for a system to be able to generalise to a variety of unseen scenarios. This is not only computationally expensive, but also compromises the transparency of the model learned, especially if very large and/or trained on cloud platforms \citep{Vasudevan2021}.

Map-based algorithms are used as a biologically realistic navigation technique, where the robotic agent acquires a global representation of the environment through a first round of exploration and then formulates plans to reach a certain target location on the basis of episodic memory, scene recognition and self-localisation in the environment. The environment is represented as a collection of observations linked by correlation relationships.
The density of the map can be variable: like in RL, each state may represent a square in a grid~\cite{Milford2004}, or a matching function~\citep{Zou2019} to establish whether an observation is new or not. Typically, the focus is on creating a topological, rather than geometrical, environment representation.

Trajectory planners simply generate trajectories for a robot to follow based on the position of obstacles or targets in the environment without requiring any training or previous exploration.
Trajectories can be calculated based on nodes in a graph, randomly generated (see \citealp{Karur2021} for an in-depth review), or tailored to sensory inputs using force fields (e.g. Vector Field Histogram \citep{Borenstein1991} or Artificial Potential Fields \citep{Khatib1985}). Again, the discretisation of the space in trajectory planning can be variable, but it typically tends to follow the grid technique, and planning occurs over steps of fixed size.
Trajectory planners are open-loop controllers in that environment feedback is not needed to generate the trajectory itself. However, closed-loop input control is used for trajectory \textit{following}. As an embodied agent is oblivious to its trajectory, this requires an external observer (e.g. a camera on the ceiling).

In its simplest form, a closed-loop controller is set up with hard-wired connections, which essentially generate one behaviour (e.g. attraction or repulsion) \textit{afforded} by a certain stimulus, without any learning required. For example, in trajectory-following, a single controller receives as input the deviation from the desired trajectory, and the output is always a steering manoeuvre in the opposite direction to that deviation. Similarly, when considering an agent such as a Braitenberg vehicle~\citep{Braitenberg1986}, the underlying control loop will always respond to a stimulus with the same control strategy (e.g. turning in the opposite direction of an obstacle). Thus, by itself, a single controller is not suitable for the execution of multi-step planning or complex behaviours. However, when considering a Braitenberg-style agent, stimuli are represented by typically transient objects: thus, a control loop can be created on demand in response to each object and disposed of once it
is no longer relevant. We call these temporary closed loop controllers \textsl{Tasks}. For this reason, Tasks are variable in their duration and scope, and do not depend on the agent's location in a global, discrete frame of reference. This opens the opportunity to create plans by chaining Tasks.

The fact that even a simple fruit fly is able to formulate complex plans in unknown environments~\citep{Honkanen2019} challenges the notion that learning through trial and error or acquiring episodic memory is necessary at all for planning. In fact,~\citet{Spelke2007} suggest that even newborn animals display a seemingly innate understanding of basic properties of the environment, such as physics and causality, which they call ``core knowledge''. An agent equipped with core knowledge can form reasonable predictions over the outcome of an action without needing to perform it, which makes for fast and efficient decision making.

We hereby present a framework in which closed-loop Tasks are combined with core knowledge to formulate plans as sequences of Tasks.
These Task sequences can be organised into a topological cognitive map
which can be stored and modified by a long-lived
supervising module~\citep{Porr2005} as fit to fulfill a certain goal. 
We call this module ``Configurator'' (cfr.~\citet{LeCun2022}).

The paper is organised as follows. 
In Section~\ref{instructional} we provide an instructional example to create an intuition for closed-loop Tasks, the use of core knowledge, and key aspects of the Configurator's role; in Section~\ref{theory} we describe in detail the components of the proposed framework; in Section~\ref{imple} we discuss the features of our specific implementation; in Section~\ref{design} we discuss the details of experimental design;
in Section~\ref{sect:results} we present the results of the experimental validation of the framework; in Section~\ref{disc} we discuss the results in light of the state of the art.

\section{Methods}

\begin{figure}[htb!]
\centering
    \includegraphics[width=0.7\textwidth]{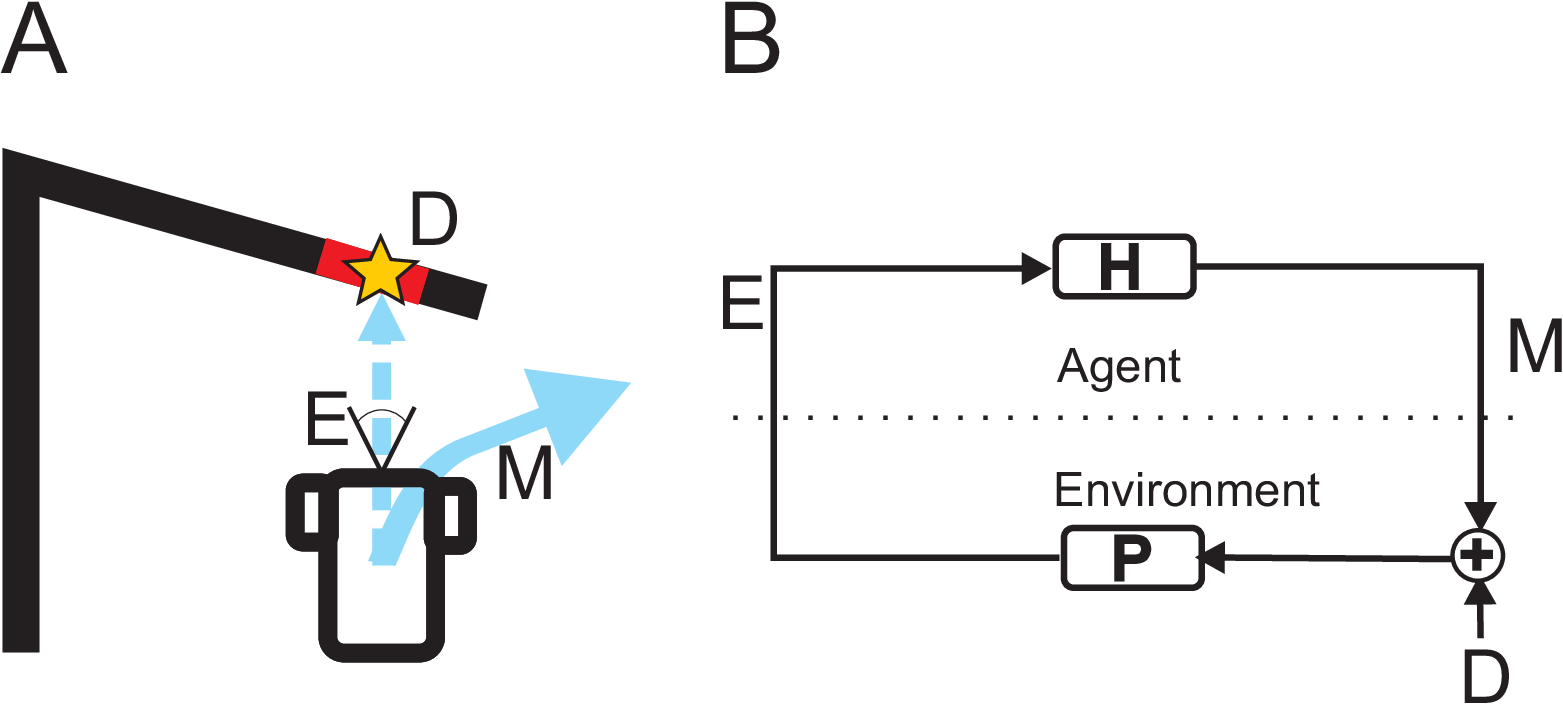}
    \caption{ \doublespacing {A: closed-loop obstacle avoidance. B: a closed-loop controller, i.e. a Task. D: disturbance (i.e. collision point), E: sensor error signal between robot and disturbance, M: motor output, P: environmental transfer function, H: transfer function of the agent/robot.}}
    \label{fig:Task}
\end{figure}

\subsection{Instructional example\label{instructional}}
In Fig.~\ref{fig:Task}A we present an informal definition of a Task. A Task is created when a disturbance (in this case obstacle $D$) enters the robot's sensor range. The sensors produce a signal $E$ which is transformed by $H$ into the reflex motor output $M$. A Task executes until the disturbance ceases to produce a signal, at which point the Task simply terminates. When no disturbance is present, the system falls back on a ``default'' Task which characterises the resting state. Importantly, unlike in classical closed-loop control, where the control problem is tackled by a single controller at all times, each Task is discarded upon termination and may be replaced by any other Task. The creation of Tasks requires a supervisory module which not only needs to create Tasks on demand but also is able to combine them. Consequently, we enhance our system by introducing the \textsl{Configurator}, a module with working memory which enables the agent to plan over sequences of Tasks.

\begin{figure}[thb!]
    \centering
    \includegraphics[width=\textwidth]{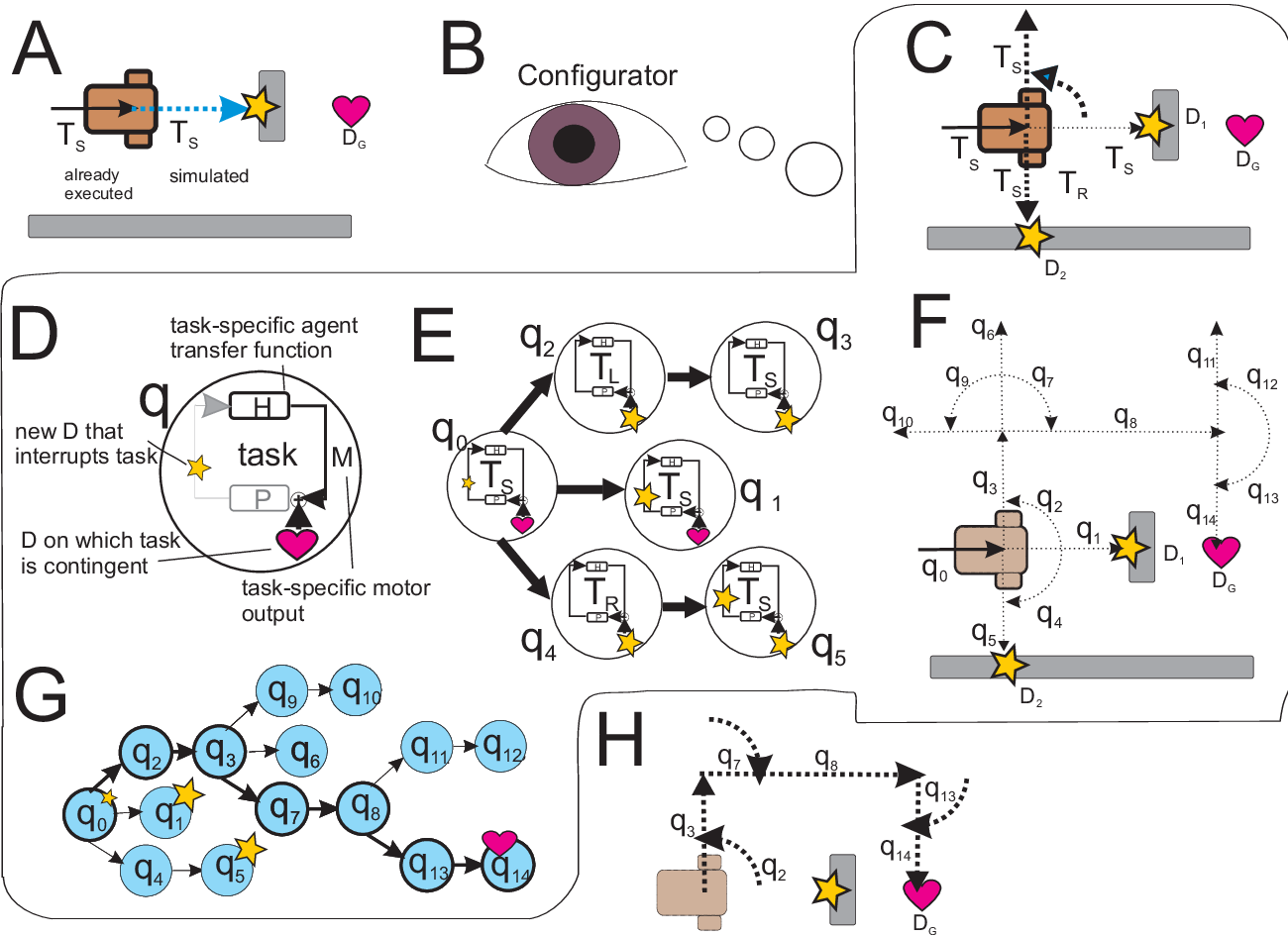}
    \caption{\doublespacing
    Example of the multi-step-ahead planning procedure carried out by the Configurator. Disturbances are indicated with stars. A: scenario requiring multi-step planning, B: the Configurator, C: possible alternatives to initial Task $T_S$ which results in a collision ($D_1$), D: a state $q$, E: cognitive map summarising the options in Fig.~\ref{fig:react_loop}C, F: trace of all the possible behaviours simulated by the Configurator, G: graph representation of Fig.~\ref{fig:react_loop}F, H: plan extracted. }    
    \label{fig:react_loop}
\end{figure}

Figure~\ref{fig:react_loop} illustrates the role of the Configurator in the planning process. Figure~\ref{fig:react_loop}A presents a scenario requiring multi-step planning: a robot is driving straight in a Task $T_S$ (solid arrow) aimed at reaching the target $D_G$ in an environment where an obstacle in the way. Obstacle $D_1$ (marked with a star) is identified \textit{before} a collision occurs by \textsl{simulating} the continuation of the present Task $T_S$ (dotted arrow). This simulation represents the agent's core knowledge. At this point, the Configurator (Fig.~\ref{fig:react_loop}B) must initiate planning. First, the currently executed Task $T_S$ is terminated (see Task solid arrow in Fig.~\ref{fig:react_loop}C). 
At this point the Configurator can explore its options to avoid obstacle $D_1$. Fig.~\ref{fig:react_loop}C shows two sets of consecutive Tasks being created and simulated to avoid obstacle $D_1$. In one set, the robot turns left (Task $T_L$) and proceeds straight (Task $T_S$), in the other the robot turns right (Task $T_R$) and then drives straight (Task $T_S$). 

The results of the simulation of each Task are stored in a searchable structure, a cognitive map, for future reference (see Fig.~\ref{fig:react_loop}E). This map represents a collection of causally linked states $q$ (see Fig.~\ref{fig:react_loop}D). States summarise a Task as a Task-specific agent transfer function $H$ which generates a Task-specific motor output, i.e. \textit{what} behaviour was performed, any initial disturbance on which the Task is contingent, i.e. \textit{why} the behaviour was needed, and any new disturbances which may interrupt Task execution, i.e. whether the behaviour was successful. The cognitive map for the simulation in Fig.~\ref{fig:react_loop}C is depicted in Fig.~\ref{fig:react_loop}E. The index number of each state denotes order of simulation.
Note that disturbance $D_1$ encountered the simulation corresponding to state $q_1$ has propagated to initial state $q_0$, as both states represents essentially different components of the same Task; however, as the robot does not collide with obstacle $D_1$ in state $q_0$, the interrupting disturbance is considered as looming, hence the smaller size in the representation. 
The full cognitive map created in this obstacle-overtaking scenario is displayed in Fig.~\ref{fig:react_loop}F. The behaviour of the robot executing the plan is shown in Fig.~\ref{fig:react_loop}H.

\subsection{Formal definitions} \label{theory}
\subsubsection{Tasks as closed-loop controllers}
Formally, we define Tasks as closed-loop controllers, as depicted in Fig.~\ref{fig:Task}B. Tasks are contingent to a disturbance $D$ which enters the Environment transfer function $P$, and generates an error signal $E$. This signal is used by the Agent (i.e. the robot) transfer function $H$ to generate a motor output $M$ aimed at counteracting the disturbance (e.g. avoiding the obstacle in Fig.~\ref{fig:Task}A). Once the disturbance is counteracted, the error signal becomes zero: at this point, the control motor output $M$ is no longer needed and the execution of the control loop terminates. 

\subsubsection{Hybrid control\label{switch}}
We model the Task-switching process using a nondeterministic hybrid automaton~\citep{Henzinger2000, Raskin2005, Lafferriere1999}. This formalism allows to capture the nature of Tasks as dynamically evolving, temporary control-loops characterised by distinct behaviours.

A hybrid automaton is a tuple \((T, F, K, I, J)\), where
\begin{itemize}
\item $T$ is a set of Tasks such that $T:\textbf{H} \times C$, where
\begin{itemize}
    \item $\textbf{H}$ 
    is a finite set of Task-specific agent transfer functions corresponding to discrete control behaviours
    \item $C$ is a set of continuous variables representing input disturbances. We use $\dot{C}$ to represent the derivatives of $C$, or the evolution of the continuous variables over time, and variables $C \ '$ to denote the valuation of $C$ with which the next control mode is initialised.     
\end{itemize}
\item \(K \subseteq H\times H\) represents the edges, or permitted transitions between control modes
\item $F, I$ are functions assigning each a predicate to each control mode. 

 \(F: \textbf{H} \rightarrow C \cup \dot{C}\) is the flow function. $F(H)$
defines the possible continuous evolutions $\dot{C}$ of the system in control mode $H$. These depend on the displacement and/or rotation generated by Task-specific motor outputs.

\(I: \textbf{H} \rightarrow C\) is the invariant function. $I(H)$
represents the possible valuations for variables $C$ in control mode $H$.

\item \(J: K \rightarrow C\) is the jump function. 
$J(k)$ is a guard which determines when the discrete control mode change represented by edge $k \in K$ is allowed based on the valuation of the continuous component $C$.
\end{itemize}

\subsubsection{Cognitive map}\label{cognitive_map}
Simulated Tasks can be summarised as states (recall Fig~\ref{fig:react_loop}D) linked by causal relationships in a cognitive map. 
We define a cognitive map as a tuple
\((Q, \hookrightarrow)\), where \(Q:H \times C\) is a set of states and \(\hookrightarrow \subseteq Q \times Q \) is a transition relation. Importantly, states $q$ differ from Tasks $T$ in that while in Tasks the focus is on their execution as control loops, in states $q$ information about the agent transfer function and disturbances serve to efficiently store the Task and its outcome.

We denote 
\begin{equation}
    Pre(q) = \{ q\,' \in Q \mid (q\,' \hookrightarrow q)\}
\end{equation}
\begin{equation}
    Post(q) = \{ q\,' \in Q \mid (q \hookrightarrow q\,')\}
\end{equation}
Further, we define a path \(\rho=q_{0}q_{1}...q_{n}\) where \(q_{i} \in Post(q_{i-1}) \forall i \geq 1\), and $q_0$ is the root state in the cognitive map. Paths represent sequences that may potentially constitute a plan.

\subsubsection{The Configurator} \label{conf_form}
We define the Configurator as a type of supervisor \citep{Ramadge1984}:
\begin{equation}
    \text{Configurator}=(\mathcal{HA}, \mathcal{G}, \Sigma, D_G, \Psi, R)
\end{equation}\label{conf_eq}
where 
\begin{itemize}
    \item $\mathcal{HA}$ is a hybrid automaton
    \item $\mathcal{G}$ is a cognitive map
    \item $D_G$ is the (possibly empty) set containing the overarching goal in the form of a disturbance which the \textsl{plan} aims to counteract
    \item $\Sigma:\mathcal{HA}\rightarrow \mathcal{G}$ is a function through which cognitive map $\mathcal{G}$ is synthesised from hybrid automaton $\mathcal{HA}$
    \item \(\Psi:\hookrightarrow \rightarrow [1,0]\) is a guard which assigns to each possible transition a supervisory control pattern \(\psi \in \Psi\), which enables (\(\psi(\hookrightarrow)=1\) or disables (\(\psi(\hookrightarrow)=0\)) transition along that edge. These assignments determine which states $q$ will be used to formulate the final plan.
    \item \(R: K \rightarrow C \cup C\,'\) is the
    reset function. $R(k)$ assigns to edge $k$ in hybrid automaton $\mathcal{HA}$ possible updates $C \ '$ for the continuous variables $C$ when a discrete change occurs. This will determine the initial continuous state at the start of a Task.
\end{itemize}

\begin{figure}[!htb]
    \centering
    \includegraphics[width=0.9\textwidth]{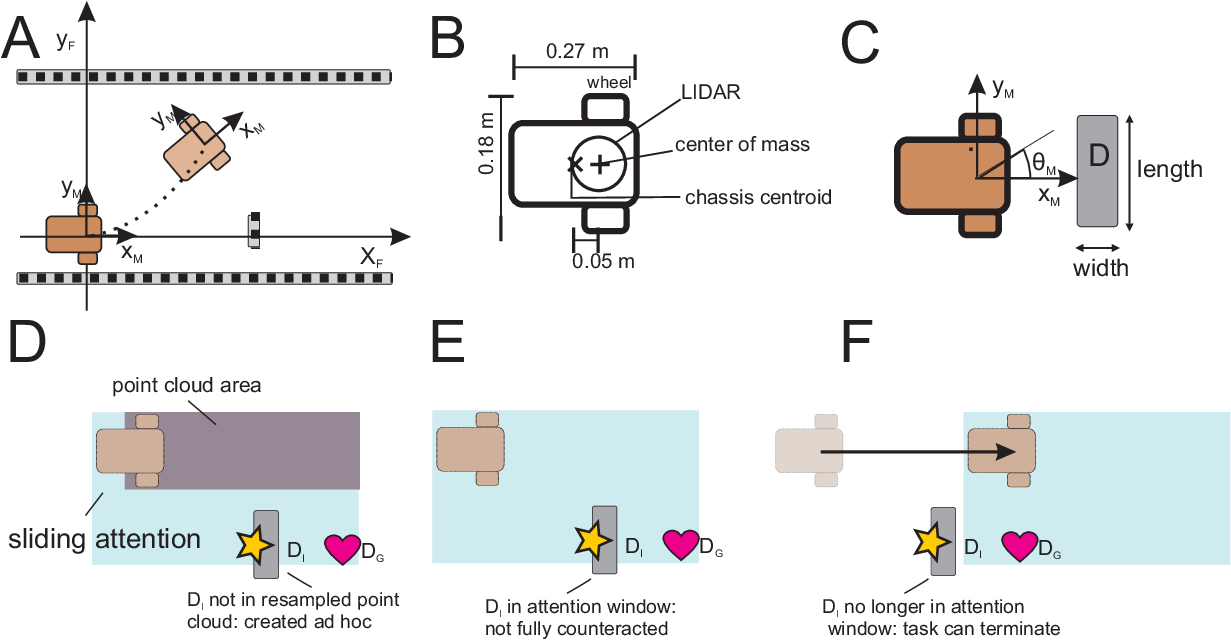}
    \caption{\doublespacing  A: frames of reference in the Box2D simulation,
     B: schematic representation of the robot with dimensions and center of mass annotated, 
     C: disturbance $D$ representation, D: definition of attention window and comparison with point cloud resampling area, E: validation of initial disturbance $D_I$ using the attention window at the beginning of a Task, F: invalidation of initial disturbance $D_I$ when it ceases to overlap with the attention window.
    }
    \label{fig:robot}
\end{figure}

\subsection{Implementation} \label{imple}
\subsubsection{Robot}
We implement the outlined framework on an indoor robot based on the Alphabot\footnote{\url{https://www.waveshare.com/alphabot-robot.htm}}. The robot is equipped with a Rasbperry Pi Model 3b+ (Broadcom BCM2837B0 processor with a 1.4~GHz clock speed and 1~GB RAM), an A1 Slamtec 2D LIDAR sensor and two 360$^o$ continuous rotation servos which move the wheels. The LIDAR callback runs at 5~Hz and the motors are updated at 10~Hz. We define the motor update rate as \(step_{motor}\). In this work, planning must satisfy the requirement of occurring within the 100~ms motor update intervals.

\subsubsection{Core knowledge as a physics simulation} \label{core}
To find a suitable multi-step plan to execute in the real world, the behaviour of the automaton in Section~\ref{switch} can be simulated in physics engine Box2D\footnote{\url{https://box2d.org/}}, which represents the system's core knowledge.

Objects in the environment are constructed from the Cartesian coordinates of the points captured by the on-board LiDAR sensor. This frame of reference is ego-centric to the real robot, and fixed (marked with axes $x_F, y_F$ in Fig.~\ref{fig:robot}A) throughout the construction of the cognitive map. We further define a moving frame of reference, indicated in Fig.~\ref{fig:robot}A with axes $x_M, y_M$, which is local to the simulated robot.
The robot was modelled in Box2D as a simple box of dimensions 0.27m$\times$ 0.18m, respectively, with the center of mass shifted forward by 0.05m from the centroid of the box (see Fig.~\ref{fig:robot}B). At the start of the simulation, the robot's center of mass is positioned at the origin of both the fixed and moving planes. As the simulation progresses, the robot’s coordinates may change in the fixed plane as it performs Tasks, while the moving plane's origin remains fixed at the robot’s center of mass. An arbitrary distance limit $r=1 \ \text{m}$ is imposed on task execution, to prevent the simulation from running indefinitely.

We define a labelling function
\begin{equation}\label{lambda}
    \lambda :Q \rightarrow V \cup V
\end{equation}
such that  $\lambda(q)=(v^{0}, v^{d})$, where vector
$v^{0}=(x^{0}, y^{0})$ represents the position of the robot in the fixed plane at the start of the Task and vector $v^{d}=(x^{d}, y^{d})$ is the displacement of the robot from the start to the end of the Task.

Simulated sensors can be added to objects in order to collect information about disturbances.
In this work we use Contact Listeners as proximal sensors reporting collisions between the simulated robot and obstacles.
Tasks are simulated one at a time, and each simulation is interrupted as soon as a collision is reported. Collision sites denote obstacles. Disturbances are represented as tuples $D=(x_M, y_M, \theta_M, w, l)$. As shown in Fig.~\ref{fig:robot}C, $x_M, y_M$ and $\theta_M$ express the position and orientation of the disturbance with respect to the simulated robot's local frame of reference, represented as the smaller Cartesian planes with axes $x_M, y_M$ in Fig.~\ref{fig:robot}A and~\ref{fig:robot}C, $w$ is its width and $l$ is its length. As shown in Fig.~\ref{fig:robot}D, if an initial disturbance $D_I$ is  not included in the point cloud, it is reconstructed \textit{ad hoc} using this information and included in the Box2D environment.

Further, we use simulated distal sensors to create a sliding attention window used to assess whether an initial disturbance $D_I$ has been fully counteracted based on whether it is or may get in the way of goal $D_G$. As shown in Fig.~\ref{fig:robot}D, this window is a box bounding the simulated robot and the goal $D_G$. The attention window is created anew at the beginning of a Task.
As shown in Fig.~\ref{fig:robot}E, at the beginning of a Task contingent upon disturbance $D_I$, if $D_I$ is included in the attention window, the disturbance is not considered fully counteracted as it may get in the way of the goal. Otherwise, it is invalidated and the counteracting Task is terminated.
The attention window is fixed to the robot and moves along with it throughout Task execution (see Fig.~\ref{fig:robot}F). Once the initial disturbance $D_I$ is no longer contained in the window, it is invalidated and the Task contingent to it terminates.

\begin{figure}[!htb]
    \centering
    \includegraphics[width=0.95\linewidth]{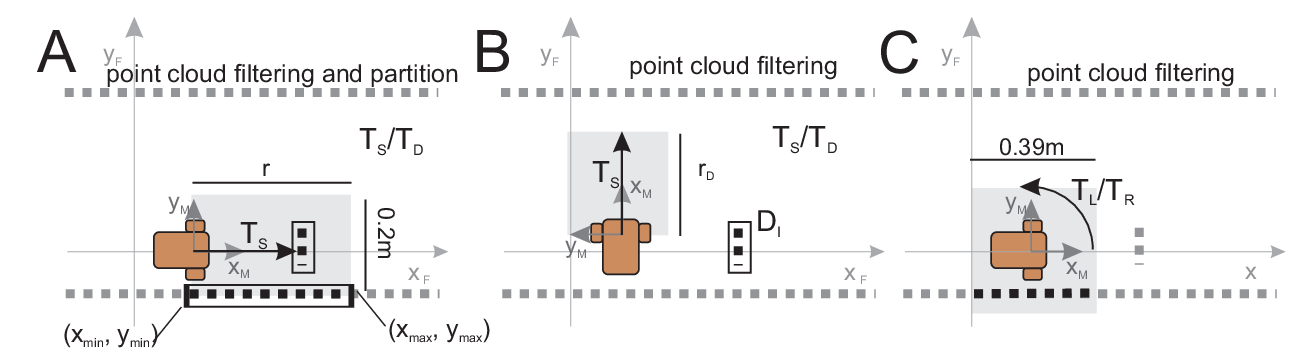}
    \caption{\doublespacing A: Point cloud filtering strategy for Tasks $T_S$ and $T_D$ with $D_I= \varnothing$ and example of partition of the resampled subset with object extraction. B: Point cloud filtering strategy for Tasks $T_S$ with $D_I \neq \varnothing$.. C: Point cloud fitlering strategy for Tasks $T_L$ and $T_R$. }
    \label{fig:resample}
\end{figure}

Here, we limit the range of the sensor data used for the simulation to $r = 1 \ \text{m}$ from the origin. This represents a planning horizon, and the simulation stops when its limit is reached. The simulation is advanced using a kinematic model of the robot with
a time step (\(step_{Box2D}= \frac{1}{10}\text{s}\)), 3 position iterations and 8 velocity iterations. 
A simulation may carry on for an arbitrary number of steps, indicated as \(n_{Box2D}\).

To reduce the number of objects represented in the simulation, first the point cloud is filtered so that only the points relevant to the Task at hand are retained. The filtering strategies for Tasks in which the robot is driving straight ($T_S$ and $T_D$) are summarised in Fig.~\ref{fig:resample}A and~\ref{fig:resample}B, and for Tasks in which the robot turns ($T_R, T_L$) on its axis in Fig.~\ref{fig:resample}C. In all figures, the points outside the shaded area (indicated in grey) are discarded; the retained points are indicated in black. The shaded area begins at the robot's center of mass and has fixed size. For Tasks $T_S$ and $T_D$, its length is 0.2$\text{m}$ and its width is equal to a task distance limit $r$ if disturbance $D_I$ is not an obstacle (Fig.~\ref{fig:resample}A), and a limit $r_{D}=x_{D}^w + w_{robot}$ otherwise (Fig.~\ref{fig:resample}B), where $x_{D}^w$ is the furthest $x$ coordinate of disturbance $D_I$ from the robot, in the robot's local frame of reference, and $w_{robot}$ is the width of the robot.
For Tasks $T_R$ and $T_L$, the shaded area is a 0.39$\text{m}\times$ 0.39 $\text{m}$ square.  Using an off-the-shelf algorithm for clustering, the resampled point cloud is broken up into clusters made up of points less than 0.1$\ \text{m}$ apart, with each cluster representing a rectangular object (see Fig.~\ref{fig:resample}A, bounding boxes around darker points). 
The extreme coordinates (in Fig.Fig.~\ref{fig:resample}A, the bottom left $(x_{min}, y_{min}) \text{and top right corner} (x_{max}, y_{max})$ corners) in each cluster are used to obtain the dimensions of the object. Respectively, the width is calculated as $\| x_{max} - x_{min} \|$ and the length as $\| y_{max} - y_{min} \|$. Its center of mass is obtained by averaging the coordinates in the cluster.

The simulation results can then be used to determine motor commands for Task execution in the real world. We calculate each motor command as a duration expressed as a number of motor update intervals \begin{equation}
    n_{motor} = n_{Box2D} \cdot step_{motor} \cdot step_{Box2D}
\end{equation}
where $step_{Box2D}$ is the rate of Box2D simulation update in seconds,
\(n_{Box2D}\) is the number of steps of duration $step_{Box2D}$ taken to bring a Task to termination in simulation and $n_{Box2D}\cdot step_{Box2D}$ calculates the simulated time taken by the Task. 

\subsubsection{Hybrid automaton} 
\begin{figure}[!htb]
    \centering
    \includegraphics[width=.9\textwidth]{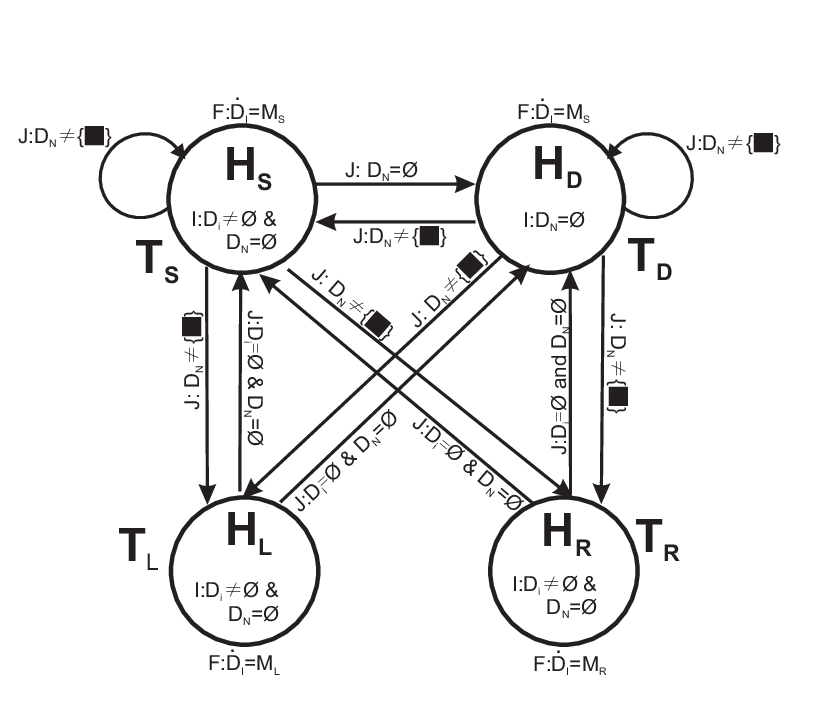}
    \caption{\doublespacing A: State diagram summarising the rules for Task-switching and execution. B: A scenario requiring hindsight state resets. C: Reset of the source state (zigzag arrow). D: Cognitive map resulting from the reset. 
    }
    \label{fig:S_hat}        
\end{figure}

The hybrid automaton modelling the simulated Task-switching process is depicted in Fig.~\ref{fig:S_hat}A. 

Agent transfer functions $\textbf{H}$ and continuous inputs $C$ are inscribed in the circles representing Tasks $T$. The arrows connecting Tasks represent the edges $K$. Flow $F$ predicates are annotated outside of each circle. Invariant $I$ predicates are annotated inside each circle. Jump $J$ predicates are located adjacent to the corresponding edges.

We define four Task types: $T_S$, indicating the behaviour of driving straight ahead in order to counteract a disturbance, $T_D$ as the ``default'' behaviour of driving ahead with no disturbance to counteract, $T_R$, indicating a right turn of the robot on its axis and $T_L$, indicating a turn to the left.
We define the agent transfer function set as $\textbf{H}={H_S, H_D, H_R, H_L}$, where $H_S$ appies to $T_S$, $H_D$ to $T_D$, $H_R$ to $T_R$ and $H_L$ to $T_L$. 
Continuous variables are defined as $C = \{ D_{I}, D_{N}\}$, where $D_{I}$ is empty or contains the disturbance that the present Task is contingent on, and $D_{N}$ is the empty set or single-element set of disturbances encountered during Task execution (similar to Fig.~\ref{fig:react_loop}D). With reference to Fig.~\ref{fig:react_loop}E, we indicate obstacles with which the (simulated) robot has collided with $\blacksquare$ , looming obstacles with $\square$ and targets with $\heartsuit$.

The flow $F$ of the system is described by velocities $M$ which are the output of the Task-specific agent transfer function.
As indicated by the corresponding invariant predicate $I$, a ``default'' Task $T_D$ executes as long as the new disturbance set $D_N$ is empty.
Tasks $T_L, T_R$ and $T_S$ execute as long as the initial disturbance has not been eliminated ($D_I\neq \varnothing$) and set $D_{N}$ is empty.
For Tasks $T_L$ or $T_R$, any initial disturbance $D_I$ is considered to have been counteracted when a turn of $\frac{\pi}{2}$ or $-\frac{\pi}{2}$ has been completed in a Task, respectively. For Tasks $T_S$, an initial disturbance $D_I=\{\heartsuit\}$ is counteracted when its x-coordinate in the robot's local frame or reference is $x_M \approx 0$. For obstacles, we define different termination criteria in Section~\ref{design} depending on the application. 

Edges are labelled with predicates of the form $J, R$, corresponding to the edge's jump and reset functions.
As indicated by the jump predicates, transitions along edges $(H_S, H_L)$, $(H_S, H_S)$, $(H_D, H_S)$, $ (H_D, H_L)$, $ (H_D, H_R)$ and $(T_S, T_R)$ are permitted if set $D_N$ does not contain any of the obstacles $\blacksquare$.
Transitions $(H_L, H_S)$, $(H_L, H_D)$, $ (H_R, H_D)$ and $(H_R, H_S)$ are permitted if set $D_I$, previously composed of one element, is empty, and if set $D_N$ does not contain new disturbances. Transition $(H_S, H_D)$ is only permitted if set $D_N$ is empty.

\subsubsection{State manipulation and reinterpretation}

To increase the agent's ability to formulate complex plans, we introduce disturbance shift (Section~\ref{shift}) and a state split (Section~\ref{split}) functions which allow for manipulation of the elements of a state $q$ in a cognitive map. In addition, to prevent unnecessary state-space explosion, in Section~\ref{phi} we define cost functions to determine priority of state expansion.

\paragraph{Disturbance shift} \label{shift}
Given disturbance $D=(x,y,\theta, w, l),$ and shift vector $v=(x^v, y^v)$, we define a $\texttt{shift}$ function :
\begin{align*}
    \texttt{shift}(D, v)=\hat{D} \text{ \ , \ where \ }
    \hat{D}=(x+x^v, y+y^v, \theta, w, l )
\end{align*}
which shifts the position of disturbance $D$ by 2D vector $v$.

\begin{figure}
    \centering
    \includegraphics[width=0.95\linewidth]{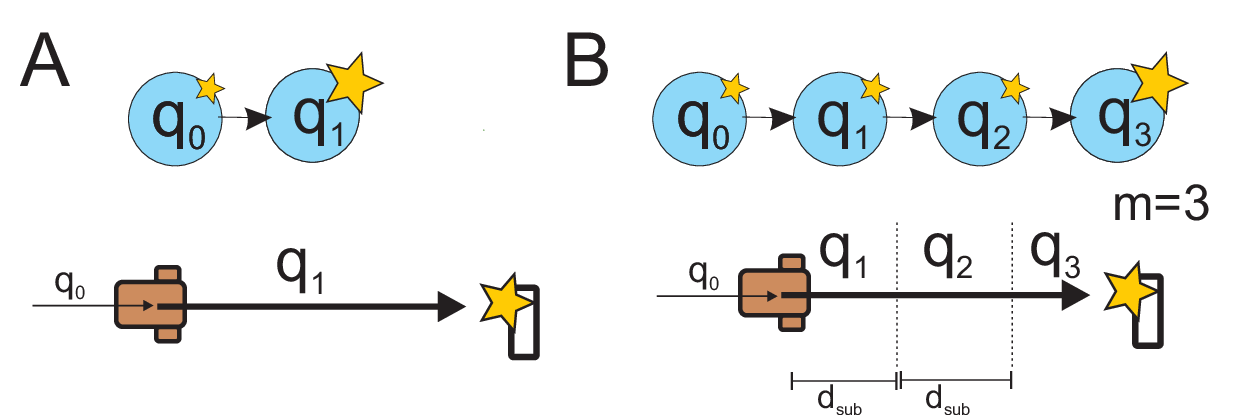}
    \caption{\doublespacing
    Example of state split. A: initial states in the cognitive map  before splitting. B: split of state $q_1$ into $m=3$ sub-states.}
    \label{fig:split}
\end{figure}

\paragraph{A posteriori state split} \label{split}
Given their flow and invariant functions (see Fig.~\ref{fig:S_hat}), the duration of Tasks $T_S$ is purely dictated by the presence of obstacles or by whether the arbitrary distance limit $r$ has been reached (see Section~\ref{core}). As shown in Fig.~\ref{fig:overtaking-pics}B, this can lead to unsuccessful planning in some more complex scenarios.
To prevent this, we give the Configurator the ability to retroactively split a state $q$ (see Fig.~\ref{fig:split}A) in the cognitive map in $m$ sub-states set a fixed distance vector $d=(x, y)$ apart, as shown in Fig.~\ref{fig:split}B.
In this way, it becomes possible for the Agent to explore the option to terminate a Task early if necessary to find a successful plan.

Formally, we define a $\text{split}$ function as follows:
\begin{equation} \label{split_eq}
\texttt{split}(q=(h, c=(D_I, D_N))
\begin{cases}
    \{ \hat{q}\ | \ \hat{q}=(h, \hat{c}=( {\hat{D}_{I}} ,{\hat{D}_{N})} ) \} & \ \text{if} \ D_N= \blacksquare \\
    \{ q \} & \ \textit{if} \ D_N = \varnothing 
\end{cases}
\end{equation}
where
\begin{align*}
    \hat{D}_I &=\texttt{shift}(D_I, v^0 + (n \cdot d_{sub})), \\
    \hat{D}_N &=\texttt{shift}(D_N, (v^0+v^d) - (n \cdot d_{sub}), \\
    (v^0, v^d) & = \lambda (q),  \\
  \forall \ n \in \{ \ m \in \mathbb{Z}^+ & \ | \ 1 \leq m \leq \lfloor \frac{(v^0+v^d)}{d_{sub}} \rfloor \}  \cup \{ \frac{(v^0+v^d)}{d_{sub}} -\lfloor \frac{(v^0+v^d)}{d_{sub}} \rfloor \}
\end{align*}    

Continuous states are indicated with a $\hat{ \ }$ symbol, $\lambda$ is a labelling function defined in Equation~\ref{lambda}, vector sum $(v^0 + v^d)$ represents the position of the robot at the end of the Task summarised in state $q$, and the notation in the form $\lfloor \frac{a}{b} \rfloor$ indicates integer division.

\paragraph{Cost-based expansion priority assignment} \label{phi}
The priority of expansion of a state $q$ is defined by its cumulative past and expected cost (cfr. \citealt{Hart1968}),
which we will define as function $\phi: Q \rightarrow \mathbb{R}$. 

Cost evaluation function $\phi$ is defined as:
\begin{equation}
\phi(q) = \gamma(q)+\chi(q)
\end{equation}
\label{ev_function} 
where, for state $q$  in cognitive map $\mathcal{G}$, $\gamma(q)$ is the past cost associated with $q$ and $\chi(q)$ is a heuristic function which estimates the cost to reach goal $D_G$ from $q$.

\begin{figure}
    \centering
    \includegraphics[width=0.6\linewidth]{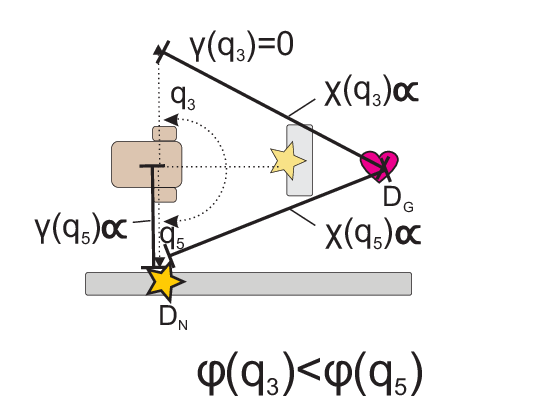}
    \caption{\doublespacing
    Calculation of past cost $\gamma$ and heuristic cost $\chi$ in two states $q_3$ and $q_5$. Function $\chi$ is directly proportional to the distance from a target $D_G$; function $\gamma$ is inversely related to the distance from an obstacle $D_N$. As state $q_3$ has a lower cumulative cost $\phi$ compared to state $q_5$, it will have priority for expansion.}
    \label{fig:phi}
\end{figure}

Fig.~\ref{fig:phi} provides a visual representation of the functions determining state expansion priority. 
In Fig.~\ref{fig:phi}A two states are represented: $q_3$, which represents a Task executed without interrupting obstacles, and $q_5$, where the robot has collided with a wall.
Functions $\gamma(q)$ and $\chi(q)$ are, respectively, related to the distance from any obstacles encountered during Task execution and from the goal (see Fig.~\ref{fig:phi}A). Specifically, as will be detailed, function $\gamma$ is related to the negative distance from the collision site. As shown in Fig.~\ref{fig:phi}B, state $q_3$ has a lower cost function, calculated as $\gamma(q_3)+\chi(q_3)$, than state $q_5$, and it is given priority of expansion.

We define the past cost function $\gamma$ as:
\begin{equation}
    \gamma(q) = \begin{cases}
         0 & \text{if} \  D_N = \varnothing \\ 
        \dfrac{| \dfrac{d_{\blacksquare} - \| \ (x^N_M \ , \ y^N_M) \ \|}{d_{max}} | + | \dfrac{\theta_{\blacksquare} - \| \ \theta^D_M \ \|}{\theta_{max}} |}{\texttt{max}_{\gamma}} & 
             \text{if}  D_N= \{\square \} \\
        \dfrac{| \dfrac{d_{\blacksquare} - \| \ (x^N_M \ , \ y^N_M) \ \|}{d_{max}} | + | \dfrac{\theta_{\blacksquare} - \| \ \theta^D_M \ \|}{\theta_{max}} | + \kappa}{\texttt{max}_{\gamma}} & 
             \text{if}  D_N= \{\blacksquare \} 
    \end{cases}
\end{equation}
where $D_N$ represents the disturbance encountered in state $q$, 
$d_{\blacksquare}= r$ and $\theta_{\blacksquare}= \frac{\pi}{2}$ are, respectively, the arbitrarily defined desired distance and orientation from an obstacle, $(x^N_M \ , \  y^N_M)$ and $\theta^N_M$ are, respectively, the position vector and the orientation of disturbance $D_N=(x^N_M, y^N_M, \theta^N_M, w, l )$ with respect to the simulated robot's moving frame of reference (see Section~\ref{core}, Fig.~\ref{fig:robot}A, F), 
\(d_{max}=2r\) and $\theta_{max}=\pi$ represent, respectively, the maximum possible distance and orientation of the robot with respect to a disturbance, $\kappa=2$ is an arbitrarily set collision penalty and $\texttt{max}_{\gamma}=6$ is the maximum value which $\gamma(q)$ can take, used for normalisation.

We define the cost heuristic function:
\begin{equation}
    \chi(q) = \begin{cases}
         0 & \text{if} \ D_G = \varnothing\\
        \dfrac{| \dfrac{d_{\heartsuit} - \| \ (x^G_M \ , \ y^G_M) \ \|}{d_{max}} | + | \dfrac{\theta_{\heartsuit} - \| \ \theta^G_M \ \|}{\theta_{max}} |}{\texttt{max}_{\chi}} & \text{if} \ D_G \neq \varnothing
    \end{cases}\
\end{equation}
where $D_G$ represents the Configurator's goal, 
$d_{\heartsuit}= 0$ and $\theta_{\heartsuit}=0$ are, respectively, the arbitrarily defined desired distance and orientation from a target, $(x^G_M \ , \  y^G_M)$ and $\theta^G_M$ are, respectively, the position vector and the orientation of disturbance $D_G=(x^G_M, y^G_M, \theta^G_M, w, l )$ with respect to the simulated robot's moving frame of reference, and $\texttt{max}_{\chi}=4$ is the maximum value possible for $\chi(q)$, used for normalisation.

\subsubsection{Synthesis of cognitive map $\mathcal{G}$} \label{synthesis}
Function $\Sigma$ is a function through which the Configurator constructs 
cognitive map $\mathcal{G}$ as a model of the surroundings. 
This representation is ego-centric to the robotic agent and provides an interpretation of sensor inputs in terms of the Tasks which they \textit{afford} the agent in the context of reaching a goal, if present.

Algorithm~\ref{alg:sigma} provides pseudocode for the hybrid automaton unwinding procedure. 

\begin{algorithm}
    \caption{\doublespacing Synthesis $\Sigma$ of cognitive map $\mathcal{G}$}
    \begin{algorithmic}
        \State Inputs: $\mathcal{HA}, \mathbf{D}_G$,
        \State initialise state-space $Q:={q_0}$, priority queue $PQ:=\{q_{0}\}$, state to expand $q_{E}:=q_{0}$, closed set $CS:= \varnothing$
        \Repeat
        \State set \(q_{E}:= \min_{q \in PQ}\phi(q)\)
        \State remove \(q_E=(h_E, c_E)\) from \(PQ\)
       \State find 
       \For{each edge $k=(h_E, h_F), \in K$}
       \Repeat
       \State{set control mode $h_F:=h_2$}
       \State{initialise Task $t:=(h_F, c_F)$, inputs $c_F:= \{ D_I=\hat{R}(k), D_N=\varnothing \}$}
       \State{initialise frontier state $q_F=(h_F, c_F)$ and add to set $Q$}
       \State{simulate Task $t$}
       \State{update disturbance $D_N$}
       \State{set edge $k:=(h_F, \cdot) \in K$} 
       \Until{control mode $h_F=(H_S, \cdot)$ or inputs $c_F=(\cdot, D_N = \{ \blacksquare \})$}
       \EndFor
        \State{find $\mathbf{q}_{F}:=\texttt{split}(q_F, \lambda(q_F))$}
        \State add $\mathbf{q}_{F}$ to $PQ$
        \Until{no more jumps are permitted from $q_{E}$ or $D_{G}$ has been reached}
    \Return \(\mathcal{G}\)
    \end{algorithmic} \label{alg:sigma}
\end{algorithm}

Reset function $\hat{R}$ will be defined in Section~\ref{design}, as it varies with the experimental design.

\subsubsection{Guard $\Psi$}
Guard $\Psi$ is used to extract a plan $\rho$ from cognitive map $\mathcal{G}$. Formally, the plan is
\begin{equation}
    \rho = \{q_0, q_1...q_n  \ | \ (q_i, q_{i+1}) \in \hookrightarrow \forall i \in \mathbb{N} \ , \  q_n= \min_{q \in Q} \phi(q)\}
\end{equation}
or a set of adjacent states where the last state has the least cost $\phi$ in the set of states $Q$. This results in the assignment of $\psi(q)=1$ if $q\in \rho$ and zero otherwise.

\section{Experimental design}\label{design}
\subsection{Experimental conditions}
We define five cognitive map-building strategies characterised by different levels of Configurator intervention in the unwinding of the hybrid automaton $\mathcal{HA}$ into cognitive map $\mathcal{G}$. 

\subsubsection{Strategy 0: reactive behaviour}\label{pilot0}
Strategy 0 is developed to test whether reactive behaviour is sufficient to fulfill the overarching goal of the Configurator. The robot does not construct a cognitive map at all, but rather generates one Task at a time in response to an immediate disturbance, i.e. it will turn away from obstacles and towards targets. Disturbance are identified by simulating only the currently executing task. Tasks of type $T_S$ and $T_D$ are executed for a fixed distance of size $d_{sub}$, to give the reactive agent the chance to produce more complex behaviour.
\subsubsection{Strategy 1: multi-step planning, ``vanilla''}
Strategy 1 serves to test the performance of multi-step ahead planning without any state manipulation by the Configurator, i.e. no state split, and without the use of the sliding attention window. The resulting map is organised topologically, meaning that states represent Tasks of variable length.

The reset function is defined in this strategy as

\begin{equation}
 \begin{aligned}
    \hat{R}(k)=
    \begin{cases}
        D \ '_I= D_G &  \ \text{if} \ D_N = \varnothing  \\
        D \ '_I= D_N &  \ \text{if} \ D_N = \{ \square \} \\
    \end{cases}
    \label{reset1-eq}  
\end{aligned} 
\end{equation}

Thus, successful Tasks are initialised with initial disturbance $D_I=D_G$. 
\subsubsection{Strategy 2: step-wise Task execution}~\label{step_wise}
Tasks $T_S$ and $T_D$ have the potential to be boundless in the cases that a) the Task is not able to eliminate initial disturbance $D_I$ or b) the Task is not interrupted by a disturbance $D_N$. To pose a boundary to the execution of these Tasks, in Strategy 2 they
are executed in step-wise increments of fixed length $d_{sub}$, each making up a state $q$ in the cognitive map. The step-wise increment is an adaptation of grid-style, geometrical discretisation of the environment in order to allow for potentially more flexible behaviour compared to Strategy 1. For example, the robot may be allowed to drive towards a known obstacle for a certain distance before initiating an avoidance maneuvre, rather than performing an avoidance manoeuvre as soon as the obstacle is detected. The reset $\hat{R}$ defined in Equation~\ref{reset1-eq} is used, meaning that the attention window is not used to assign initial disturbances $D_I$.
\subsubsection{Strategy 3: state split only}\label{exp3}
In Strategy 3, we build upon Strategy 1 by introducing the  the $\texttt{split}$ function defined in Section~\ref{split}, as an alternative means to increase behaviour flexibility while only increasing the size of the state-space $Q$ when needed. 
States are broken down into sub-states only if they terminate in a collision, thus potentially leading to a sparser, more topologically organised cognitive map compared to Strategy 2. As the attention window is not introduced, reset $\hat{R}$ is used here. 
\subsubsection{Strategy 4: state split + sliding attention window}
\label{exp5}
Strategy 4 represents the full expression of the cognitive map-building strategy detailed in Algorithm~\ref{alg:sigma}. Here, states are, like in Strategy 3, split retroactively only as needed, and, thanks to the use of the sliding attention window, the entire plan can be tailored to the size and position of obstacles and targets, while maintaining a topological, rather than geometrical, organisation of the cognitive map. The Configurator uses reset
\begin{equation}
\begin{aligned}\label{reset2}
    R(k) = \begin{cases}
         D\,'_I=D_G & \text{if}  \ \  \texttt{in\_view}(D_I) = \texttt{false} \wedge D_N=\varnothing\\
        D \ '_I= D_N &  \ \text{if} \ \ D_N = \{ \square \}   \\
        D\,'_I=D_I & \text{if} \ \  \texttt{in\_view}(D_I)= \texttt{true} \wedge D_N=\varnothing 
    \end{cases}\
\end{aligned}    
\end{equation}

where $D\,'_I$ is the set of initial disturbances with which the next task will be initialised, and function $\texttt{in\_view}$ returns true if initial disturbance $D_I$ and the robot's attention window overlap (recall Section~\ref{core} and Figg.~\ref{fig:robot}D,~\ref{fig:robot}E and~\ref{fig:robot}F).

\subsection{Experimental scenarios}
Two experimental scenarios were defined. In the first and simplest, the robot is placed in front of a cul-de-sac which it must simply avoid. In the second scenario, we test simultaneous target pursuit and obstacle avoidance in a bounded race track environment, where the robot must overtake an obstacle in order to reach a target. 
The performance of the robot was evaluated with three different starting points with respect to the cul-de-sac and three different starting points with respect to the obstacle in the race track. Three experimental runs were carried out per starting position, with a total of nine experimental runs per Strategy in each scenario.
\subsection{Data analysis}
Across these sets of experiments, we first evaluate the ability to formulate plans able to fulfil a certain overarching goal (e.g. obstacle avoidance or reaching a target). Secondly, we measure the time taken to construct cognitive map $\mathcal{G}$ and formulate a plan, the size of the state-space $Q$, and number of Box2D objects (as this number has a major influence on the time complexity of the simulation).
Kruskal-Wallis tests were used to assess the presence of between-group differences in the total number of Box2D objects represented, the number of states making up the cognitive map in each experimental run, and the time taken to build the state-space and forumulate a resulting plan. Where a significant test result was reported, post-hoc Mann-Whitney tests were carried out to uncover the source of these differences.
Finally, we explored the Pearson correlations among time performance, number of Box2D objects and size of the cognitive map in the entire dataset.
\section{Results}\label{sect:results}
\subsection{Cul-de-sac avoidance}
In this scenario, the robot is simply set in front of a cul-de-sac which it must avoid in order to avoid collisions. The Configurator's goal is set as $D_G=\varnothing$, and the fixed Task split distance was set to $d_{sub}=0.5 \text{m}$.
\begin{figure}
    \centering
    \includegraphics[width=\linewidth]{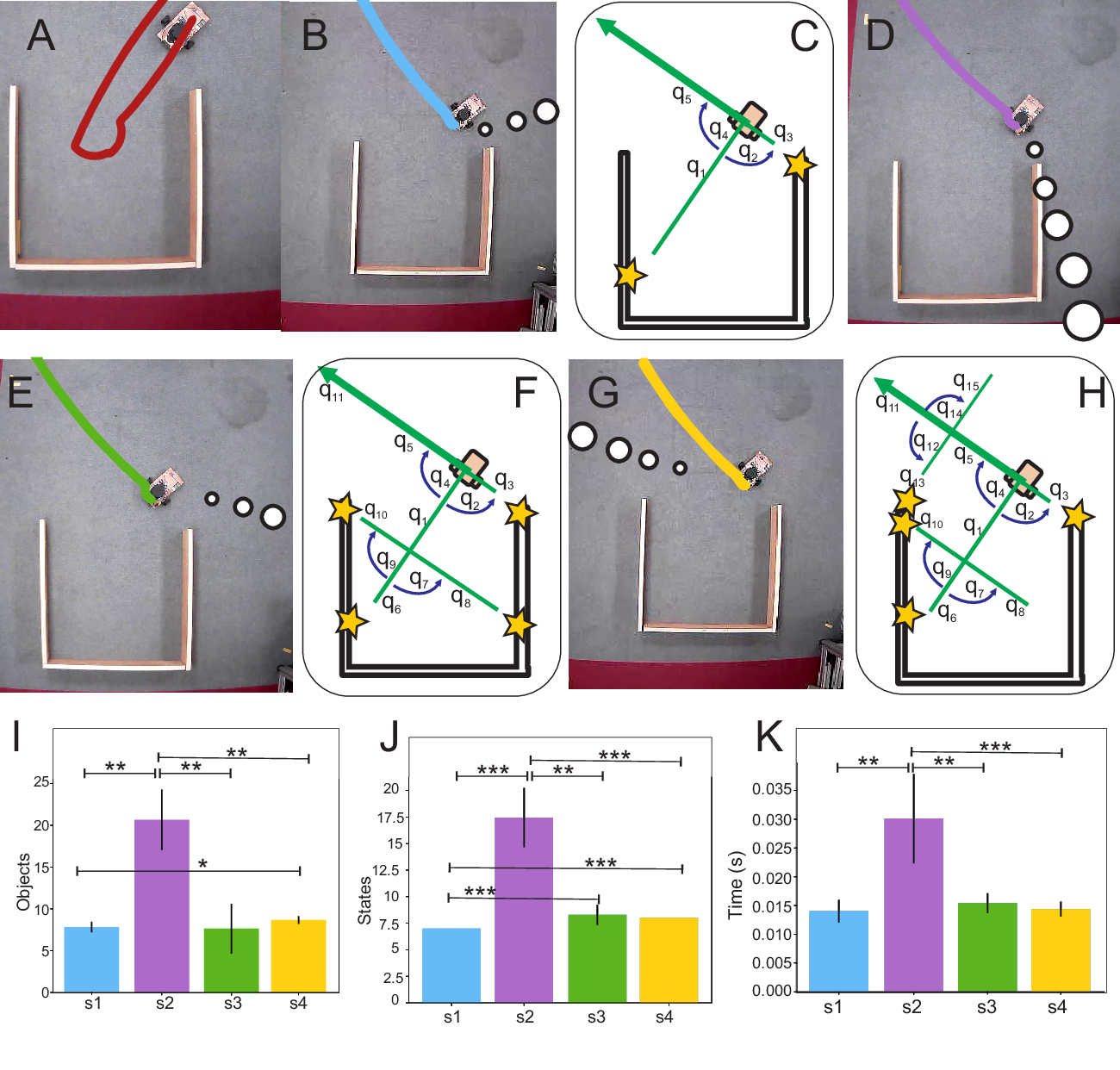}
    \caption{\doublespacing Results of cul-de-sac avoidance experiments. A: with strategy 0 (reactive behaviour), B: with strategy 1 (Task chains, no split), C: cognitive map for Strategy 1, D: with Strategy 2 (Task chains, step-wise execution), E: with strategy 3 (Task chains, state split), F: cognitive map for strategy 3 and 4, G: with strategy 4 (Task chains with Task split and sliding attention window), H: cognitive map for strategy 2, I: statistics for object measurements, J: statistics for state measurements, K: statistics for time measurements. }
    \label{fig:cds}
\end{figure}

Fig.~\ref{fig:cds} depicts the performance of the robot in the five tested implementations. Figg.~\ref{fig:cds}A, \ref{fig:cds}B, \ref{fig:cds}D, \ref{fig:cds}E and \ref{fig:cds}G depict the behaviour exhibited by the robot when using decision-making Strategy 0, 1, 2, 3 and 4, respectively. As can be observed, a reactive approach results in inefficient avoidance of the cul-de-sac, where the robot enters the cul-de-sac first and then turns away from it once it comes to a distance $d_{sub}$ from the walls. On the other hand, enabling the agent to plan over sequences of closed-loop behaviours always produced a successful plan, with identical behaviour across experimental runs. Despite the similarities in behaviour, as Figg.~\ref{fig:cds}C, \ref{fig:cds}F and \ref{fig:cds}H show, the construction of the state-space differed between strategies. 

\begin{table}[]
    \centering
    \begin{tabular}{c|c|c|c}
    \hline
        {}&\textbf{$\mu$ Objects} & \textbf{$\mu$ States} & \textbf{$\mu$ Time (s)}  \\
         \textbf{Strategy 1}&7.78 $\pm$ 0.63& 7.0 $\pm$ 0 & 0.0139 $\pm$0.002 \\
        \textbf{Strategy 2}&20.67 $\pm$ 3.62& 17.44 $\pm$ 2.78 & 0.030 $\pm$0.008 \\
        \textbf{Strategy 3}&7.67 $\pm$ 2.98& 8.33 $\pm$ 0.94 & 0.0153 $\pm$0.002 \\
        \textbf{Strategy 4}&8.67 $\pm$ 0.47& 8.0 $\pm$ 0 & 0.014 $\pm$0.001 \\
    \end{tabular}
    \caption{\doublespacing Descriptive statistics for strategies 1, 2, 3 and 4.}
    \label{tab:descr_cds}
\end{table}

Table~\ref{tab:descr_cds} reports the descriptive statistics for strategies 1, 2, 3 and 4, while inferential statistics are depicted in Fig.~\ref{fig:cds}I, \ref{fig:cds}J and \ref{fig:cds}K. Strategy 0 was excluded from inferential statistical analyses as no state-space was constructed. Planning over closed-loop behaviours always satisfied the real-time requirement of occurring within 100~ms. Significant between-groups differences were found in the number of objects simulated ($u=27.03, p<0.001$), states constructed ($U=33.69, p<0.001$) and time for cognitive map construction and planning ($U=22.09, p<0.001$). For post-hoc comparisons the significance cutoff $\alpha$ was set as 0.0083 to adjust for multiple comparisons.
The most simulated objects represented was found in Strategy 2, and the difference was significant compared to Strategy 1 ($U=81.0, p<0.001$), 3 ($U=0.0, p<0.001$) and 4 ($U=81.0, p<0.001$)  . In Strategy 3 significantly fewer objects were simulated compared to Strategy 4 ($U=9.0, p<0.0083$), but the differences between Strategy 1 and Strategy 3, and Strategy 4 and Strategy 4 were not significant.
The cognitive map built in Strategy 2 had the highest number of states (Fig.~\ref{fig:cds}), which was significantly higher than Strategy 1 ($U=81.0, p<0.0001$), Strategy 3 ($U=81.0, p<0.001$) and Strategy 4 ($U=81.0, p<0.0001$). Although Strategy 1 resulted in around one less state in the cognitive map compared to Strategy 3 and 4, this difference was statistically significant (for both comparisons, $U=0.0, p<0.0001$). 
Following a similar pattern, Strategy 2 resulted in a significantly slower performance compared to all other strategies (compared to Strategy 1: $U=81, p<0.001$; compared to Strategy 3: $U=0.0, p<0.001$; compared to Strategy 4: $U=81.0, p<0.0001$).

\subsection{Obstacle overtaking}
\begin{figure}
    \centering
    \includegraphics[width=\linewidth]{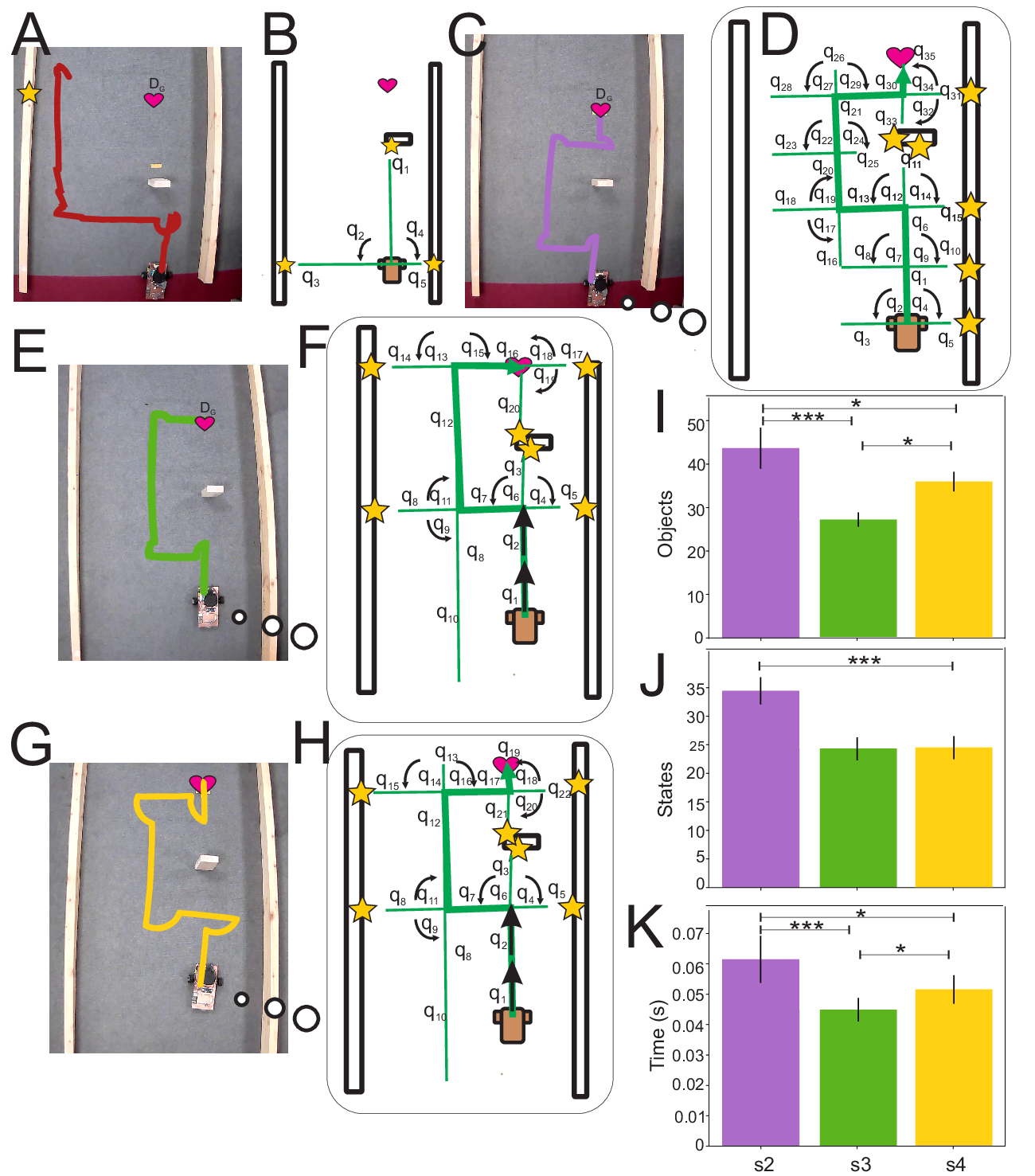}
    \caption{\doublespacing Results of obstacle-overtaking experiments. A: behaviour of reactive agent (Strategy 0), B: state-space obtained using cognitive map building Strategy 1, C: behaviour of agent using Strategy 2, D: cognitive map constructed with Strategy 2, E: behaviour of agent using Strategy 3 in the race track, F: cognitive map constructed with Strategy 3, g: behaviour of agent using Strategy 4, H: cognitive map constructed with Strategy 4, I: statistics for objects measurements, J: statistics for state measurements, K: statistics for time measurements.}
    \label{fig:overtaking-pics}
\end{figure}

In the ostacle overtaking experiments, the robot is instructed to reach a target $D_G=(x=1.0m, y=0.0m)$ in a cluttered race track. In this experiment, we set split size $d_{sub}=.27~\mathrm{m}$  As anticipated, and as shown in Figure~\ref{fig:overtaking-pics}A, reactive behaviour (Strategy 0) is not sufficient to reach the target in this scenario. Additionally, while decision-making Strategy 1 was sufficient to avoid the cul-de-sac, this was not the case for this more complex scenario. In fact, no plan could be formulated, as the world representation is limited to seven states, all of which resulted in collision (Fig.~\ref{fig:overtaking-pics}B).

As shown in Fig.~\ref{fig:overtaking-pics}C,  \ref{fig:overtaking-pics}E, \ref{fig:overtaking-pics}G,
using any among Strategies 2, 3 and 4, the Configurator was able to find a plan to reach target $D_G$, with possible slight variations in behaviour motivated by the different underlying definitions of Task termination criteria and construction of the state-space (Fig.~\ref{fig:overtaking-pics}D, \ref{fig:overtaking-pics}F, \ref{fig:overtaking-pics}H, ). In Fig.~\ref{fig:overtaking-pics}C and~\ref{fig:overtaking-pics}D, where the robot executes Tasks in steps of distance $d_{sub}$ to build its cognitive map (Strategy 2), after the obstacle is overtaken, the robot quickly returns in line with the target, as state $q_30$ has the lowest $\phi$ in the frontier. On the other hand, in Strategy 3 (Fig.~\ref{fig:overtaking-pics}E, \ref{fig:overtaking-pics}F), the robot first drives as close as possible to the target, and then to the target position. With Strategy 4 (Fig.~\ref{fig:overtaking-pics}G and  \ref{fig:overtaking-pics}H), the robot again returns in line with the target immediately after the overtaking manoeuvre is completed, similarly to the behaviour exhibited by the robot in Strategy 2, but with a different underlying cognitive map. Specifically, the construction of the state-space in Fig.~\ref{fig:overtaking-pics}D resembles a grid-like representation, while the granularity of the cognitive map in Fig.~\ref{fig:overtaking-pics}H varies with the duration of the Tasks simulated. It is worth noting that, while successful plans were formulated by the robot using strategies 2, 3 and 4, Strategy 3 only yields a successful plan because state $q_7$ terminates in collision and is therefore split into $q_7$ and $q_8$. If the race track had had no bounding walls, and no artificial distance limits $r$ were posed to Task execution, planning would have been unsuccessful, as the simulation of the Tasks corresponding to states $q_5$ and $q_7$ would have continued for infinite time. This demonstrates the benefit and necessity of the addition of a sliding attention window in Strategy 4.

\begin{table}[]
    \centering
    \begin{tabular}{c|c|c|c}
    \hline
        {}&\textbf{$\mu$ Objects} & \textbf{$\mu$ States} & \textbf{$\mu$ Time (s)}  \\
         \textbf{Strategy 2} &43.67 $\pm$ 4.71&  34.44 $\pm$ 2.36 & 0.062 $\pm$0.015 \\
         \textbf{Strategy 3}&27 $\pm$ 1.63& 24.33 $\pm$ 2.06 & 0.045 $\pm$0.004 \\
        \textbf{Strategy 4}&35.89 $\pm$ 2.23& 24.44 $\pm$ 2.01 & 0.051 $\pm$0.005 \\
    \end{tabular}
    \caption{\doublespacing Descriptive statistics for strategies 2, 3 and 4.}
    \label{tab:descr_overtaking}
\end{table}
Table~\ref{tab:descr_overtaking} displays descriptive statistics for strategies 2, 3 and 4. Strategies 0 and 1 were excluded from the analyses as they did not yield successful plans. For post-hoc analyses, we set significance cutoff $\alpha=0.0167$ to adjust for multiple comparisons. Significant differences were found across experimental runs in the number of objects represented in the Box2D simulation ($H=22.229, p<0.001$),  in the size of the state-space($H=24.137, p<0.001$), and in time to cognitive map construction and planning ($H=21.082, p<0.001$).
The results of post-hoc analyses for measurements pertaining to objects are depicted in Fig.~\ref{fig:overtaking-pics}I. The number of bodies represented in Box2D in Strategy 2 was significantly higher than in Strategies 3 ($U=0, p<0.00167$) and Strategy 4 ($U=7.5, p<0.0167$). Additionally, in Strategy 4, significantly more Box2D objects were simulated compared to Strategy 3 ($U=0, p<0.0167$).  Strategy 2 also resulted in a significantly larger state-space compared to Strategy 3 and to Strategy 4(for both, $U=0, p<0.000167$, also see Fig.~\ref{fig:overtaking-pics}J). The inferential statistics for comparisons between planning times across experiments are displayed in Fig.~\ref{fig:overtaking-pics}K. Strategy 2 yielded the highest time taken for cognitive map construction and planning, and this was significantly higher than that in Strategy 3 ($U=2.0, p<0.00167$), and Strategy 2 ($U=8.0, p<0.0167$). However, Strategy 4 had significantly higher planning time compared to Strategy 3 ($U=10.0, p<0.0167$).

\subsection{Correlations}
\begin{figure}[!htb]
    \centering
    \includegraphics[width=\linewidth]{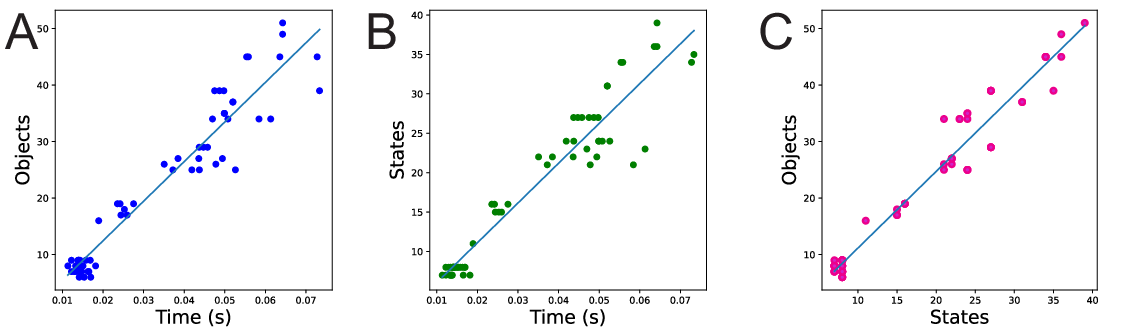}
    \caption{\doublespacing Scatter plots illustrating the relationships between A: planning time and Box2D objects, B: planning time and state-space size, C: Box2D objects and state-space size.}
    \label{fig:correl}
\end{figure}

Very strong correlations were found between time performance and number of objects represented in the Box2D simulation ($R=0.96, p<0.001$, Fig.~\ref{fig:correl}A), between time performance and state-space size ($R=0.959, p<0.001$, Fig.~\ref{fig:correl}B) and between number of number of states in the cognitive map and Box2D objects ($R= 0.978, p<0.001$, Fig.~\ref{fig:correl}C). 

\section{Discussion\label{disc}}
We have introduced a novel technique to achieve multi-step planning using pure input control. Behaviours are here defined as closed-loop, object-centered Tasks, and a Configurator module was designed which is able simulate and evaluate sequences of Tasks in a physics engine which represents the system's ``core knowledge''. This approach has the advantages of being entirely landmark- (and not trajectory-) based, fully online, designed for low-power robots and unseen environments, and focused on multi-step reasoning over a sparse cognitive map, rather than on policy selection over a dense state-space. Taken together, these aspects make for a transparent and trustworthy technique suitable for deployment in a variety of automated vehicles.

The planning problem involves constructing a sequence able to fulfil a goal; intimately tied with this is the problem of how to discretise the environment. An even grid, however fine, is a popular solution for trajectory-based techniques~\citep{Khatib1985, Hart1968}. More recently, and similarly to the approach taken in this work, the problem has been structures as a hybrid one, where a discrete behaviour ``skeleton'' is constructed, and continuous, behaviour-specific, control functions are synthesised. In these classes of problems, discrete behaviours are viewed as abstract actions. A major challenge is therefore to determine the bounds of the discrete behaviour in the continuous environment.

One solution is to apply space- or time constraints to discrete behaviours. Space constraints simply represent the start and end point of a trajectory, and trajectories can be sampled from the environment and optionally optimised~\citep{Garrett2018, Kim2022, Thomas2021}, or synthesised from scratch~\citep{Migimatsu2020}. Time constraints can further be applied: the behaviour-specific control functions is any one which generates trajectories between two points in a limited time~\citet{Constantinou2018, Mavridis2019}. These methods scale well to object manipulation problems, where each behaviour represents an object-oriented action (e.g. pick up or move a box), albeit with relatively slow planning times, in the range of seconds~\citep{Migimatsu2020, Kim2022, Garrett2018}. However, in spatial navigation, it is possible that sampling-based approaches may perform suboptimally in planning long routes in cluttered environments without sub-goals posing further spatial constraints to the final trajectory (cfr.~\citet{Constantinou2018, Mavridis2019}). Further, trajectory planning has drawbacks discussed in Section~\ref{intro}.

Another approach to finding behaviour boundary is through learning. This is typically done in an ``option'' framework for RL, where each option is an abstract behaviour with a distinct initiation and termination state, and an action distribution~\citep{Salter2022, Wulfmeier2020}. In this framework, a high-level controller is learned to switch between low-level options, whose boundaries are also learned. State-based approaches generally are well-suited to embodied robots, and the introduction of an option/skill/Task level which is hierarchically higher to actions improves sample efficiency. The key difference between the option framework and our own is that the former is essentially a problem of segmentation of trajectories in an experience dataset. This paradigm for deep RL learns options boundaries and transitions in a dense state-space and in hindsight, which makes it unsuitable for fully online applications. On the other hand, in our proposed method, Task boundaries can be quickly and deterministically identified, based on the dimensions of objects in the surroundings, thanks to the use of the attention window. This results in a reduced size of the constructed state-space and a reduced need for random exploration, which make our algorithm suitable for immediate and safe deployment on a naive agent. Additionally, the small state-space and definite Task initiation and termination criteria allow for framework transparency and interpretabily, which are aspects in which deep RL falls short~\citep{Chen2024}.

The use of a gaming engine to explore the state-space is not, \textit{per se} novel: rather, gaming engines are widely used, in substitution to or alongside real-world data, to generate synthetic data for RL model training, e.g.~\citet{Anwar2019, Osinski2020, Ogum2024}. 
The notion that an innate knowledge base drastically cuts training time and resources is not novel either. In meta-learning~\citep{Luo2021, Bing2023}
and transfer-learning~\citep{Sun2023, Anwar2019, Salter2022} algorithms, pre-trained RL models are used as a foundation for fine-tuning a novel model. In recent years, physics-informed neural networks have been introduced~\citep{Pateras2023}, where the data-driven (offline) training of a naive model is integrated with an \textit{a priori} physics model of a particular process. In navigation, these have been used to model the fluid dynamics of pedestrian flow~\citep{Guo2024}, to precisely track the location of a vehicle via multi-sensory integration~\citep{Imbiriba2024}, and to model vehicle and obstacle dynamics in order to generate a collision-free trajectory~\citep{Sanyal2024}.
In this work, the physics simulation is not used to generate realistic training data, but to explore the environment in terms of disturbances located in it, and generate a realistic expectation of their properties (e.g. location, shape, whether they are obstacles) and of finite Tasks by them afforded.
This allows the robotic agent to actively reason about its environment in real-time, rather than simply generalise from previous experience. This improves the reliability of our framework compared to RL, where the behaviour of the agent in unseen scenarios is unpredictable~\citep{Chen2024}.

Simulations are also used in system verification to confirm that a system satisfies certain properties. In model-checking, the behaviour of a system, summarised as an automaton, is unwound through time in its corresponding transition system. A simulation in this context is a graph search in which the properties of the system are evaluated. Model checking was applied to a real-world spatial planning problem in~\citet{Chandler2023}. In their work, planning took around $10.1~\mathrm{ms}$ over a transition system of fixed cardinality (14 discrete states) and depth (five states). Our method (Strategy 4) reports a comparable average planning time of $14~\mathrm{ms}$ with the same maximum model depth.
Verification of hybrid system is not always a decidable problem due to state-space explosion, especially due to the continuous state component~\citep{Clarke2018}. As mentioned, a solution to this problem is the use of timed automata, such as in~\cite{Constantinou2018, Mavridis2019}. Strategy 2 was partially inspired to this workaround. While the time constraint placed on each state-specific trajectory improves the quality of the solution trajectory, it does not guarantee optimality. On the other hand, our use of a cost heuristic heuristic in state-space expansion and planning ensures that the most optimal solution is always chosen.


The use of a cost heuristic to build the cognitive map was inspired by~\citet{Ulrich2000}, who developed an algorithm where an A*-style heuristic is used for navigation in partially observable environments. We estimate that the real-time performance of their Vector Field Histogram* algorithm should translate to a maximum of less than 12ms and an average of less than 2ms on the hardware used in our robot (assuming that the algorithm runs on one core only). Similarly, the response time of other path-planning algorithms tends to be at least one order of magnitude smaller than in this approach (e.g.~\citet{Amiryan2015}).
The difference in performance is very likely to be due, first of all, to the formulation of the problem as one of a hybrid nature, and the consequent need to simulate the evolution of the continuous state of the system. Secondly, our robot agent is capable of encoding and interpreting its surroundings as sequences of closed-loop behaviours contingent to particular disturbances, rather than just reacting to spatial frequency information. For this reason, additional computation time is required to carry out passes (resets) through the cognitive map.

Cognitive maps are a popular tool in robotic planning ~\citep{Epstein2017}. Like in RL paradigms, cognitive maps are organised as states representing observations of the environment at different points in time, connected by edges representing the relationship between observations. Unlike in RL, the state-space of a cognitive map can be searched and multi-step plans can be extracted as sequences of states~\citep{Tang2018}, as opposed to one policy at a time. Cognitive maps can have a sparser state-space compared to RL models. This is because RL models are often trained end-to-end, and the problem of navigating to a specific goal is entangled with the optimal control problem (e.g. ensuring that each action is executed correctly~\citep{Chen2024}). While cognitive maps can be also used for optimal control~\citep{Amirkhani2020, Mendonca2012, Motlagh2012}, they can be used in navigation for a robot to orient itself in the environment and reach a target location. This orientation and path-finding process involves the fusion of visual data with global path integration and odometry information~\citep{Milford2004, Tang2018, Zou2019}, as well as visual template matching.
In such a framework, localisation and odometry are carried out by attractor networks which need to be trained; moreover, visual information is necessarily collected through a first round of physical exploration of the environment. Plans cannot therefore be formulated in an unknown environment. Through the employment of the Box2D simulation as a means to explore the properties of objects in the environment, meaningful cognitive maps can be encoded on-the-fly by a naive robot before it initiates a behaviour, and plans can be formulated in a completely unknown environment.

As constructed by~\cite{Milford2004, Tang2018, Zou2019}, cognitive maps are heavily inspired by biology. Before being implemented in robotic navigation, they were conceptualised by~\citet{Tolman1948} to explain behaviours which could not be reduced to being a result of mere conditioning~\citep{Epstein2017}. Indeed, recall of visited locations, environment surveying~\citep{Pezzulo2014} and execution of motor plans~\citep{Verwey2014} involves sequential activation of areas related to a route or a multi-step motor plan.
Information related to episodic memory of routes is coded in hippocampal place cells~\citep{Morris2023}. 
In the discussed cognitive/experience map literature~\citep{Tang2018, Yu2023}, route planning occurs in a known environment, and the establishment of new connections between neurons is more evocative of sequential place cells activation patterns which are observed at rest, in wakefulness and sleep. These firing patterns are associated more with reward processing, map consolidation/optimisation and establishment of connectivity between experiences; thus, they have little to do with real-time, on-the-fly planning.
On the other hand, there is evidence that in an unfamiliar environment, place cells are involved in "vicarious trial and error". In vicarious trial and error, place cells coding for past, current and future locations are activated in sequences \textit{at decision time} as a means to find an efficient goal-directed plan. Along with decision-making, these firing patterns are thought to help establish event (i.e. Task) boundaries, and to chunk experiences into behaviorally significant sequences
~\citep{Pezzulo2019}. In our work, both of these functions are carried out via the Box2D simulation, which thus models the deliberative behaviour displayed by living organisms in decision-making problems in novel environments.

Place cells are known to exhibit location-specific firing patterns, and the place-specificity results from a rearrangement of the place fields called ``remapping'', which occurs within minutes of exposure to a novel environment in area CA1 of the hippocampus, but quickly vanishes when the goal or the environment change~\citep{Fenton2024, Morris2023, Ambrogioni2023}. 
Given the hippocampus' great number of in- and outgoing projections, the sequential activation of place areas in vicarious trial and error has brain-wide effects. Firstly, the rapidly emerging, but fleeting, place fields in area CA1 in the hippocampus are thought to influence the slower, but more stable, emergence of place fields in area CA3~\citep{Dong2021}, suggesting that the acquisition of a consolidated, cross-trial stable cognitive map is not necessary for successful planning on-the-fly. Moreover, at the level of the more plastic CA1 area, thanks to these neurons' disparate connections, it is possible to form fluid, constantly updating representations of experiences, called engrams~\citep{Lopez2024, Dorst2024, Miry2021}. While in traditional cognitive map-based approaches to robotic planning experiences are represented predominantly as sensor data, experimental evidence supports a link between scene recognition and affordance recall~\citep{Harel2022}, suggesting that the action performed in association with a particular scene is not a trivial aspect but a core component of a memory. In this work, each hybrid state may be regarded as a high-level description of a memory associated to an engram composed of both sensory inputs (i.e. disturbances) and behaviours (i.e. Task-specific agent transfer functions).

Indeed, the association of a stimulus with goal-directed behaviour~\citep{Katzman2020, Ho2022} reveal that interaction with the environment and choice increases the likelihood that an object will be encoded in memory. Conversely, input disturbances (e.g. objects) and goals greatly influence the segmentation of an experience into behaviorally meaningful components with defined beginning and end~\citep{Zacks2007} dictated by the manipulation afforded by each specific object~\citep{Pomp2024}. This is not the first work in which experiences (real or potential) are captured as an object and an associated action~\citep{Kruger2011}. 
Our framework provides a tool for object/affordance-based segmentation of an environment and estimation of temporal and causal relationship between closed-loop Tasks. For this reason, it may provide a solid foundation to explore the acquisition of a stable, biologically realistic map of the environment which could be used by an embodied agent to adapt previously acquired knowledge to novel scenarios.
These development will be covered in future work.


The framework presented in this paper is not without limitations. First, we designed a very small control mode space $H$ composed of simple motor commands (either driving straight or turning sharply left or right). This was a choice made to simplify the navigation problem, and dictated by the low-power robotic platform used. Complex control modes require accurate self-motion tracking, which is nontrivial in an embodied agent (which, as mentioned in Section~\ref{intro}, is oblivious to its trajectory). We designed the present application as a multi-threaded program in a four-core low-power microcontroller where one thread was assigned to the main program, one to the LiDAR sensor, one to the motors and one to the Configurator. The introduction of a new sensor (e.g. a camera) for sensor fusion would imply the introduction of a new thread, which would then determine hyperthreading and may result in degradation in speed performance. For this reason, we chose to avoid control modes requiring fine motor control to relieve the computational burden on the low-power robot used.
It is expected that increasing the size of the agent transfer function $H$ space would result in increase in time complexity of our algorithm. One simple solution to improve the smoothness of robot motion could be to code control modes $H_L$ and $H_R$ as abstract turns which may be proportional to
to measures of disturbance ``intensity'' (e.g. its size or proximity~\citep{Braitenberg1986}). This implementation may be the object of future work, although, as mentioned, this would require additional measures to ensure accurate Task execution.

Another limitation is the very strong correlation between time taken for cognitive map construction and planning and amount of Box2D objects and states. This poses concerns about scalability of the model to large, cluttered environments, such as trafficked city roads or junctions. From our experimental results, we observed that the use of a custom (Strategy 4), rather than fixed (Strategy 2), state-space discretisation resulted in a more efficient environment representation and fewer simulated Box2D objects. However, as the simulation of each Task requires the representation at least of the robot as a Box2D object, it is not possible to completely extricate decision-making time from environment complexity in our framework. There are, however, improvements that could be made to benefit scalability. For example, to avoid unnecessary state-space expansion, a more elaborate $\texttt{split}$ function could be designed to split a state into fewer sub-states, whose spacing reflects waypoints in the environment (e.g. points of reduced spatial frequency~\citep{Ulrich2000}). Additionally, the Box2D simulation step $step_{Box2D}$ and simulation range $r$ could be dynamically adapted to the complexity of the environment and goals, such that driving ahead for a long distance may be simulated in only a few simulation steps, while fine manoeuvres in a cluttered environment (e.g. reverse parking) may be allowed to simulated more precisely, with a smaller simulation step. We believe that, as is, our algorithm may be suitable for indoor service robots; the aforementioned improvements may be a good starting point to broaden the scope of real-world applications of the framework proposed in this work.

Lastly, we structured simulated Tasks as truly closed-loop behaviours, while the Tasks physically executed by the robot were based on motor commands derived from simulation results, as opposed to LiDAR inputs only. Again, this was a strategy adopted to avoid placing the burden of multi-sensory integration on the Raspberry Pi, especially in turning Tasks, where LiDAR data presents a high level of distortion due to noise. We believe that an additional strength of our framework is the fact that states constitute essentially instructions for the robot, which contain all the information necessary to perform a Task without need for external input. This application, which would greatly improve the reliability and self-sufficiency of our framework in embodied robots, may be possible even without sophisticated multi-sensory integration and should
be a focus for future work.

\section{Acknowledgements}
This work was supported by a grant from the UKRI Engineering and Physical Sciences Research Council Doctoral Training Partnership award [EP/T517896/1-312561-05]; the UKRI Strategic Priorities Fund to the UKRI Research Node on Trustworthy Autonomous Systems Governance and Regulation [EP/V026607/1, 2020-2024]; and the UKRI Centre for Doctoral Training in Socially Intelligent Artificial Agents [EP/S02266X/1].

\typeout{}
\bibliographystyle{plainnat}
\bibliography{main}
\end{document}